\documentclass[11pt,draftcls,onecolumn]{IEEEtran}
\usepackage{ifpdf}
  \usepackage{cite}
\ifCLASSINFOpdf
   \usepackage[pdftex]{graphicx}
  \DeclareGraphicsExtensions{.pdf,.jpg,.png, .eps}
\else
   \usepackage[dvips]{graphicx}
  \DeclareGraphicsExtensions{.eps}
\fi
\usepackage[cmex10]{amsmath}
\usepackage{amsthm}
\usepackage{amsfonts}
 \usepackage{algorithmic}

\usepackage{array}

\usepackage{mdwmath}
\usepackage{mdwtab}

\usepackage{eqparbox}

\ifCLASSOPTIONcompsoc
	\usepackage[tight,normalsize,sf,SF]{subfigure}
\else
	\usepackage[tight,footnotesize]{subfigure}
\fi

\ifCLASSOPTIONcompsoc
  \usepackage[caption=false,font=normalsize,labelfont=sf,textfont=sf]{subfig}
\else
  \usepackage[caption=false,font=footnotesize]{subfig}
\fi
\ifCLASSOPTIONcaptionsoff
  \usepackage[nomarkers]{endfloat}
 \let\MYoriglatexcaption\caption
 \renewcommand{\caption}[2][\relax]{\MYoriglatexcaption[#2]{#2}}
\fi
\usepackage{url}

\begin{document}
%
\title{Image Similarity Using Sparse Representation and Compression Distance}
%

\author{Tanaya~Guha,~\IEEEmembership{Student Member,~IEEE,}
        and~Rabab~K~Ward,~\IEEEmembership{Fellow,~IEEE}
\IEEEcompsocitemizethanks{\IEEEcompsocthanksitem The authors are with the Image and Signal Processing Laboratory, department
of Electrical and Computer Engineering, the University of British Columbia, Vancouver,
BC, Canada.\protect\\
\IEEEcompsocthanksitem e-mail: tanaya@ece.ubc.ca, rababw@ece.ubc.ca}
\thanks{}}

%
%

\markboth{Journal Draft,~Vol.~x, No.~x, Apr~2013}%
{Shell \MakeLowercase{\textit{et al.}}: Bare Demo of IEEEtran.cls for Computer Society Journals}
%

\IEEEcompsoctitleabstractindextext{%
\begin{abstract}
A new line of research uses compression methods to measure the similarity between signals. Two signals are considered similar if one can be compressed significantly when the information of the other is known. The existing compression-based similarity methods, although successful in the discrete one dimensional domain, do not work well in the context of images. This paper proposes a sparse representation-based approach to encode the information content of an image using information from the other image, and uses the compactness (sparsity) of the representation as a measure of its compressibility (how much can the image be compressed) with respect to the other image. The more sparse the representation of an image, the better it can be compressed and the more it is similar to the other image. The efficacy of the proposed measure is demonstrated through the high accuracies achieved in image clustering, retrieval and classification. 
\end{abstract}

\begin{keywords}
Image similarity, Compression, Kolmogorov complexity, Overcomplete dictionary, Sparse representation
\end{keywords}}

\maketitle

\IEEEdisplaynotcompsoctitleabstractindextext

%
\IEEEpeerreviewmaketitle
%
%
\section{Introduction}
\label{sec:intro}
\IEEEPARstart{M}easuring the similarity between a pair of images is of critical importance to many multimedia information processing systems involving retrieval, enhancement, copy detection, quality assessment, clustering and classification. Given the long history of image similarity evaluation, the volume of literature on this topic is large and diverse. Widely used similarity measures such as the Euclidean distance, the Mean Squared Error and other norm-based measures work well in specific cases, but they are often criticized for not corresponding well with our visual perception of similarity \cite{girod}. Another popular approach to describe the visual content of images is to extract a set of meaningful features. The similarity between two images is then computed in terms of the similarity between their features. However, the success of this approach is limited by the availability, selection and extraction of a good set of meaningful features, demanding specific knowledge of the application and the data.

Recently, there has been an interest in developing image similarity measures using \emph{compression} methods \cite{thesimilarity, Clustercompress, prdc, finitecontext, ck}. In this approach, two signals are considered similar if one can be compressed significantly when the information of the other is provided. The advantages of these methods are that they are \emph{parameter-free} (the only choice the user has to make is which compression algorithm to use) and \emph{generic} (they assume no prior knowledge of the application, and can be applied, without modification, to a variety of problems).

The compression-based similarity methods rely on a new mathematical theory of similarity which is in turn based on the idea of the \emph{Kolmogorov complexity} \cite{thesimilarity, Clustercompress}. The Kolmogorov complexity (also known as the algorithmic entropy) is a theoretical measure of randomness of a given data, and in general, is a non-computable quantity. In practice, it is often approximated by the \emph{length of the compressed data}. Intuitively, the more a given data can be compressed, the lower is its complexity. 

The compression-based similarity measures have been shown to be highly effective in clustering and classifying discrete, uni-dimensional data such as text and protein sequences \cite{thesimilarity, Clustercompress}; but their successful application in the context of real-valued, higher-dimensional data like images is scarce. For effectively measuring the similarity between two signals, the compressor being employed needs to satisfy certain properties so as to be a \emph{normal} compressor \cite{Clustercompress}. However, most state-of-the-art compressors for images (such as JPEG, JPEG2000) are \emph{not} normal, and the normal compressors (compressors of the Lempel-Ziv family) do not work well on images \cite{finitecontext}. Existing methods \cite{ndd,klz78,modeldcc} transform images into strings in order to take advantage of the normal data compressors, and thus lose the important spatial information. Another serious obstacle lies in evaluating and approximating the \emph{conditional compression} (a quantity that measures how much can a given data be compressed w.r.t.\ another data) which is the key component in every compression-based similarity measure. 

In this paper, we propose a sparse representation-based approach to encode the information content of an image; and use the compactness (sparsity) of the representation of the image as a measure of its compressibility i.e.\ how much can the image be compressed. The more sparse the representation of an image, the better it can be compressed. 

In order to design a similarity measure that correlates well with the human perception, we learn a set of basis elements (collectively called a \emph{dictionary}) from the images. This approach empowers us to build a \emph{cortex-like representation} of an image. In 1996, Olshausen and Field have shown that the basis elements that resemble the properties of the receptive field of simple cells in the primary visual cortex can be learnt from input images \cite{nature}. The keys to building such a cortex-like dictionary are: (i) a \emph{sparsity prior} - an assumption that it is possible to describe the input image using a small number of basis elements, and (ii) \emph{overcompleteness} - the number of basis elements in the dictionary is greater than the vector space spanned by the input vectors. Given a pair of images, our method learns a dictionary for each image and computes how sparsely can one image be approximated using the dictionary extracted from the other, with a required precision. 

The rest of the paper is organized as follows. Section \ref{sec:background} briefly describes the related work on compression-based distances, section \ref{sec:proposed} proposes the sparse representation-based distance measure, and section \ref{sec:experiments} presents experimental results. Section \ref{sec:conclusion} concludes the article with possible directions to future work.
\section{Previous Work}
\label{sec:background}
The work of Kolmogorov and others \cite{kolmogorov1965,solomonoff1964,chaitin1966} on how to measure data complexity has been influential in many areas of knowledge, across multiple disciplines. The notion of complexity of a string is related to its randomness. For example, the binary string $1101010001$ is considered more complex compared to the string $0101010101$, because the latter contains a regularity (repeating pattern) and therefore is less random. Kolmogorov complexity formalizes this concept. 

Given a finite object, such as a binary string $\mathbf{x}$, its \emph{Kolmogorov complexity} $K(\mathbf{x})$ is defined as the length of the shortest program that can effectively produce $\mathbf{x}$ on a universal computer, such as a Turing machine \cite{kolmogorovBook}. The Kolmogorov complexity, however, is non-computable in general. In practice, it is often approximated by the length or the file size of the compressed data. Intuitively, the more a given data can be compressed, the lower is its complexity.
\subsection{Compression-based distance measures}
\label{subsec:sparse}
Recently, Kolmogorv's theory of complexity has been used to address the problem of similarity measurement \cite{thesimilarity, Clustercompress}. Given two signals $\mathbf{x}$ and $\mathbf{y}$, a distance metric, known as the \emph{Normalized Information Distance} (NID) is developed using $K(\mathbf{x})$ and the conditional Kolmogorov complexity $K(\mathbf{x}|\mathbf{y})$. $K(\mathbf{x}|\mathbf{y})$ is defined as the length of the shortest program used by a universal computer to generate $\mathbf{x}$ when $\mathbf{y}$ is known. 

Due to the non-computable nature of the Kolmogorov complexity, a practical analog of the NID metric is proposed based on standard compression methods. This is called the \emph{Normalized Compression Distance} (NCD). Intuitively, NCD considers $\mathbf{x}$ and $\mathbf{y}$ to be similar if one can be significantly compressed when the information of the other is provided. It is defined as follows:
\begin{equation}
	\mathrm{NCD}(\mathbf{x},\mathbf{y}) = \frac{\mathrm{max}\left\{C(\mathbf{x}|\mathbf{y}), C(\mathbf{y}|\mathbf{x})\right\}}{\mathrm{max}\left\{C(\mathbf{x}),C(\mathbf{y})\right\}}
	\label{eq:ncd}
\end{equation}
The conditional compression $C(\mathbf{x}|\mathbf{y})$ is approximated as follows:
\begin{equation}
C(\mathbf{x}|\mathbf{y})=C(\mathbf{xy}) - C(\mathbf{y})
\end{equation}
where $C(\mathbf{xy})$ denotes the compressed length of the concatenation of $\mathbf{x}$ and $\mathbf{y}$. 

The NCD metric has been shown to be effective in clustering mitochondrial genomes, languages and music \cite{Clustercompress}. Following the success of NCD, different versions of compression-based distance measures have been proposed; for example, a \emph{Compression-based Dissimilarity Measure} (CDM) is proposed in the context of parameter-free data mining and is shown to be useful for anomaly detection, clustering and classification of text, DNA and time-series data \cite{parameterfree}. CDM is defined as
\begin{equation}
	\mathrm{CDM}(\mathbf{x},\mathbf{y}) = \frac{C(\mathbf{x}\mathbf{y})}{C(\mathbf{x})+C(\mathbf{y})}
	\label{eq:parameterfree}
\end{equation}
Other applications of compression-based distances include symbolic music clustering \cite{music} and plagiarism detection \cite{plagiarism}. The idea of compression, independent from NCD, has also been used to design a pattern representation scheme for automatic categorization of music, voice, genome, etc.\ \cite{prdc}; but this method requires encoding media data input into text.
\subsection{Compression-based distances for images}
\label{subsec:sparse}
In the context of images, however, successful application of the compression-based distance measures is scarce. We identify two major reasons behind that.
\begin{itemize}
	\item The success of the compression-based distances heavily depends on the availability of a \emph{normal} compressor. A compressor is normal only if it satisfies certain conditions such as idempotency, monotonicity, symmetry, etc.\ (please refer to \cite{Clustercompress} for details). The problem is that most state-of-the-art image compressors (such as JPEG, JPEG2000) are \emph{not} normal, and the normal compressors (such as the compressors of the Lempel-Ziv family) do not work well on images \cite{finitecontext}. 
	\item Another serious obstacle lies in evaluating and approximating the conditional complexity terms such as $C(\mathbf{x}|\mathbf{y})$ in NCD. These terms are the key components in a compression-based measure. The existing compression-based methods (whether or not they involve images) either approximate the conditional compression $C(\mathbf{x}|\mathbf{y})$ with $C(\mathbf{xy}) - C(\mathbf{y})$ or use a simplified definition so as not to include any conditional term (as in \eqref{eq:parameterfree}). Direct evaluation of $C(\mathbf{x}|\mathbf{y})$ is usually bypassed mainly to retain the simplicity of the compression-based measures since evaluating $C(\mathbf{x}|\mathbf{y})$ accurately requires delving into the complicated standards and algorithms of data or image compression. This also makes the compression-based methods difficult to improve upon.
\end{itemize}	 

Clearly, the straightforward extension of the methods that work perfectly well on discrete, one-dimensional data has not been very promising in the context of images. In the pursuit of alternatives, a new image encoder is proposed based on the finite context model and preliminary results on a face database are provided \cite{finitecontext}. Another recent approach, namely the CK-1 method, uses the MPEG1 video compressor to measure image similarity \cite{ck}. This method takes advantage of the temporal redundancy reduction step in video compression which performs inter-frame block matching. In this approach, a two-frame video consisting of the images to be compared is created. One frame is compressed with reference to the other frame using a standard video compressor. The compressed file size of the video is used to approximate the closeness between the pair of images. This method has been shown to be useful in texture classification.
\section{The Proposed Approach}
\label{sec:proposed}
A natural way of measuring the similarity between two given images is to quantify how well either image can be represented using the information of the other. The more similar the images, the better is the representation of one image in terms of the other. Our method formalizes this intuitive idea of similarity using a sparse representation-based approach.
\subsection{Sparsity as a measure of data complexity} 
\label{subsec:complexity}
It is well-known that sparsity of representation plays a key role in achieving good compression. For example, the superiority of JPEG2000 is mainly attributed to the capability of the wavelet transform toward representing an image more sparsely than the DCT used in JPEG. Intuitively, the more sparse the representation of a signal is, the fewer are the components needed to capture the signal's information content and the better it can be compressed. 

Sparsity thus can be seen as a direct measure of the randomness or complexity of the data. A natural image usually exhibits many repeated structures which can be discovered through its decomposition over a set of properly chosen basis functions. Due to the presence of redundancy, only a few basis functions are required to capture the significant information content of such images, resulting in a sparse representation. In the case where such structures are rare (e.g.\ in random Gaussian noise), there is no way to represent the data using a small number of basis elements. This indicates that as the complexity of a signal increases, more and more components are needed to represent the signal with a desired accuracy i.e.\ its sparsity decreases in the transform domain. This inherent connection between sparsity and data complexity is exploited in our proposed distance measure.
\subsection{Sparse Representation-based distance measure}
\label{subsec:sparse}
The basic idea in sparse signal analysis is to represent a signal by a linear combination of a small number of basis functions. Consider a signal $\mathbf{b}\in\mathbb{R}^m$ represented as a linear combination of $n$ basis functions or atoms,
\begin{equation}
\mathbf{b} = \mathcal{D} \mathbf{a}
\end{equation}
where the dictionary $\mathcal{D} \in \mathbb{R}^{m\times n}$ and its columns are the basis functions or atoms. If the values of the majority of components in $\mathbf{a}\in\mathbb{R}^n$ are $0$ (or close to $0$), we say that $\mathbf{x}$ has a \emph{sparse} representation w.r.t.\ $\mathcal{D}$. For orthogonal bases like Fourier, $\mathcal{D}$ is a square matrix i.e.\ $m=n$. For those cases where the number of basis vectors is greater than the dimensionality of the input signal i.e.\ where $m<n$, $\mathcal{D}$ is said to be \emph{overcomplete}. An overcomplete dictionary offers greater flexibility in representing the essential structures in a signal, which in turn leads to higher sparsity in the transform domain. Such representation also has advantages such as robustness to additive noise and occlusion \cite{lewicki}.\\

\subsubsection{learning the dictionries} Let us consider an image $X$. A set of $k$ random, possibly overlapping patches (each of dimension $\sqrt{m}\times \sqrt{m}$) is extracted from $X$. Every patch is converted to a vector of length $m$ and the patches are the concatenated to form a matrix $\mathbf{B}_\mathrm{x}\in\mathbb{R}^{m\times k}$. In order to build a perceptually meaningful model for $X$, we intend to learn an overcomplete dictionary $\mathcal{D}_\mathrm{x} \in \mathbb{R}^{m\times n}$ that has $n$ atoms ($m<n$) using the local patches in $\mathbf{B}_\mathrm{x}$ as input. However, greater difficulties arise with a set of overcomplete bases. An overcomplete dictionary matrix creates an underdetermined system of linear equations having an infinite number of solutions. Knowing that the natural signals are sparsely representable, often in such cases, we seek the sparsest solution i.e.\ we want the vector $\mathbf{a}$ to contain as few non-zero elements as possible. 

Our objective is to learn $\mathcal{D}_\mathrm{x}$ such that each patch (column) $\mathbf{b}_\mathrm{x_i}\in \mathbf{B}_\mathrm{x}$ can be closely approximated as a linear superposition of a small number of atoms in $\mathcal{D}_\mathrm{x}$. This is achieved by solving the following sparse optimization problem:
\begin{equation}
	\begin{aligned}
	& \underset{\left\{\mathcal{D}_\mathrm{x}, \mathbf{a}_\mathrm{x}\right\}}{\mathrm{min}}
	& & \sum_i\left\Vert\mathbf{a}_\mathrm{x_i}\right\Vert _{p}
	& \mathrm{s.t.}
	& & \forall i,\,\left\Vert\mathbf{b}_\mathrm{x_i}-\mathcal{D}_\mathrm{x}\mathbf{a}_\mathrm{x_i}\right\Vert _{2}\leq \epsilon
	\end{aligned}
	\label{eq:p01x}
\end{equation}
where the vector $\mathbf{a}_\mathrm{x_i}\in\mathbb{R}^{n}$ is the sparse representation of the patch $\mathbf{b}_\mathrm{x_i} \in\mathbb{R}^{n}$. The sparse representation of $\mathbf{B_x}$ w.r.t.\ $\mathcal{D}_\mathrm{x}$ is denoted as the matrix $\mathbf{A}_\mathrm{x} = [\mathbf{a}_\mathrm{x_1} | \mathbf{a}_\mathrm{x_2}|...|\mathbf{a}_\mathrm{x_k}]$. The value of $p$ is typically $0$ or $1$ and $\epsilon$ denotes the reconstruction error controlled by the user. 

Note that, with $p = 0$ (the $\ell_0$ seminorm that counts the number of non-zero elements in a vector) equation \eqref{eq:p01x} becomes non-convex, and solving it exactly is an NP hard problem. Approximate solution is found instead using either greedy algorithms \cite{omp} or using convex relaxation \cite{bp}. The convex relaxation methods use $p=1$ (the $\ell_1$ norm) to transform \eqref{eq:p01x} into a convex problem. 

We employ a fast dictionary learning algorithm called the K-SVD algorithm \cite{ksvd1} which provides an approximate solution to \eqref{eq:p01x} for the $\ell_0$ case. It performs two steps at every iteration: (i) sparse coding and (ii) dictionary update. In the first step, the dictionary $\mathcal{D}_\mathrm{x}$ is fixed and $\mathbf{a}_\mathrm{x_i}$ is computed by a greedy algorithm called \emph{Orthogonal Matching Pursuit} (OMP) \cite{omp}. Next, the atoms of $\mathcal{D}_\mathrm{x}$ are updated sequentially, allowing the relevant coefficients in $\mathbf{a}_{\mathrm{x}}$ to change as well. For the details of this algorithm, please refer to the original K-SVD paper \cite{ksvd1}.\\

\subsubsection{Sparse representation-based complexity functions} We define two quantities that measure the compressibility (how much can an image be compressed) of an image by (i) using its own dictionary, and (ii) using the dictionary extracted from the other image, $Y$. We name these terms as the \emph{Sparse complexity} and the \emph{Relative sparse complexity}, respectively.\\

\noindent\textbf{Definition 1.} \emph{Given an image $X$, its Sparse Complexity $S\left(X, \mathcal{D}_\mathrm{x}\right)$ is defined as the sparsity of $\mathbf{A}_\mathrm{x}$ averaged over the number of columns in $\mathbf{A}_\mathrm{x}$ i.e.\ }
\begin{equation}
S(X, \mathcal{D}_\mathrm{x}) = \frac{1}{k}\left\|\mathbf{A}_\mathrm{x}\right\|_p = \frac{1}{k}\sum_{i=1}^k\left\|\mathbf{a}_\mathrm{x_i}\right\|_p
\label{eq:sparseC}
\end{equation}
\noindent Therefore, for $p=0$, $S\left(X, \mathcal{D}_\mathrm{x}\right)$ is the average number of non-zero coefficients required to reconstruct a column of $\mathbf{B}_\mathrm{x}$ using $\mathcal{D}_\mathrm{x}$, up to a required precision $\epsilon$. Smaller value of $S(X, \mathcal{D}_\mathrm{x})$ indicates higher compressibility (lower complexity) of $X$.\\

\noindent \textbf{Properties of $\mathbf{S(X,\mathcal{D}_\mathrm{x})}$}:
\begin{itemize}
\item $S(X,\mathcal{D}_\mathrm{x})>0$ for non-empty $X$, and is  equal to $0$ otherwise.
\item Considering that $X$ is represented by $\mathbf{A}_\mathrm{x}$ and hence $XX$ is represented by $[\mathbf{A}_\mathrm{x}|\mathbf{A}_\mathrm{x}]$, we have\\
 $S(XX,\mathcal{D}_\mathrm{x}) = S(X,\mathcal{D}_\mathrm{x})$.\\
This property (idempotency) follows from the averaging operation and indicates that the sparse complexity function can compress the duplicate entries.
%
\end{itemize}

Given another image $Y$, the compression-based measures attempts to approximate how much can the image $X$ be compressed when additional information about $Y$ is available. As discussed before, this conditional quantity, is difficult to approximate and that limits the success of these measures. We hence define a slightly different complexity term that measures \emph{how much information about $X$ is contained in $Y$}. We name this term as the \emph{Relative Sparse Complexity}, .

Let $\mathcal{D}_\mathrm{y}\in\mathbb{R}^{m\times n}$ be the dictionary pertaining to the image $Y$ learnt in the same manner as $\mathcal{D}_\mathrm{x}$ (refer to\eqref{eq:p01x}). The image $X$ can be approximated in terms of the dictionary of $Y$ as follows:
\begin{equation}
	\begin{aligned}
	& \underset{\mathbf{a}_\mathrm{x|y}}{\mathrm{min}}
	& & \sum_{i=1}^k\left\|\mathrm{\mathbf{a}_{x|y_i}}\right\| _{p}
	& \mathrm{s.t.}
	\left\|\mathrm{\mathbf{b}_{x_i}}-\mathcal{D}_\mathrm{y}\mathrm{\mathbf{a}_{x|y_i}}\right\| _{2}\leq \epsilon 
	\end{aligned}
\label{eq:cond1}
\end{equation}
where $\mathbf{a}_\mathrm{x|y_i}\in\mathbb{R}^n$ is the sparse representation of $\mathbf{b}_\mathrm{x_i}$ w.r.t.\ $\mathcal{D}_\mathrm{Y}$ and $\mathbf{A}_\mathrm{x|y} = [\mathbf{a}_\mathrm{x|y_1}|\mathbf{a}_\mathrm{x|y_2}|...|\mathbf{a}_\mathrm{x|y_k}]$ is the sparse representation of $\mathbf{B}_\mathrm{x}$ w.r.t.\ $\mathcal{D}_\mathrm{y}$.\\

\noindent\textbf{Definition 2.} \emph{Given two images $X$ and $Y$, the Relative Sparse Complexity $S(X,\mathcal{D}_\mathrm{y})$ is defined as the sparsity of $\mathbf{A}_\mathrm{x|y}$ averaged over the number of columns in $\mathbf{A}_\mathrm{x|y}$.}
\begin{equation}
S(X,\mathcal{D}_\mathrm{y}) = \frac{1}{k}\left\|\mathbf{A}_\mathrm{x|y}\right\|_p = \frac{1}{k} \sum_{i=1}^k\left\|\mathbf{a}_\mathrm{x|y_i}\right\|_p
\label{eq:CsparseC}
\end{equation}
\noindent
Therefore, for $p=0$, $S(X,\mathcal{D}_\mathrm{y})$ becomes the average number of non-zero coefficients required to reconstruct a column of $\mathbf{B}_\mathrm{x}$ using $\mathcal{D}_\mathrm{y}$, up to a required precision $\epsilon$. A smaller value of $S\left(X,\mathcal{D}_\mathrm{y}\right)$ indicates that $X$ is efficiently represented by the information extracted from $Y$ i.e.\ $X$ and $Y$ have higher similarity.\\

\noindent \textbf{Properties of $\mathbf{S\left(X,\mathcal{D}_\mathrm{y}\right)}$}:
\begin{itemize}
\item $S(X,\mathcal{D}_\mathrm{y})>0$ for non-empty $Y$, and $0$ otherwise.
\item $S(XY,\mathcal{D}_\mathrm{y})= S(YX,\mathcal{D}_\mathrm{y})$ (symmetry)
\item $S(X,\mathcal{D}_\mathrm{y})>S(X,\mathcal{D}_\mathrm{x})$ for $X\neq Y$. This is because, in general, $X$ is expected to be more efficiently (sparsely) approximated using $\mathcal{D}_\mathrm{x}$ - the dictionary trained on itself, than $\mathcal{D}_\mathrm{y}$ - a dictionary trained on a different image.
\end{itemize}

\subsubsection{The distance measure} Based on the two terms defined above, a sparse representation-based distance measure $d_S$ is defined as follows:
\begin{equation}
d_S(X,Y) = \frac{S(X,\mathcal{D}_\mathrm{y}) + S(Y,\mathcal{D}_\mathrm{x})}{S\left(X, \mathcal{D}_\mathrm{x}\right) + S\left(Y, \mathcal{D}_\mathrm{y}\right)} -1
\label{eq:sdm}
\end{equation}
The proposed form of $d_S$ is much similar to that of the compression-based CK-1 distance measure \cite{ck}. From the property of the relative sparse complexity we have 
\begin{equation*}
	S(X,\mathcal{D}_\mathrm{y})>S(X,\mathcal{D}_\mathrm{x})\,\ \mathrm{and}\,\ S(Y,\mathcal{D}_\mathrm{x})>S(Y,\mathcal{D}_\mathrm{y})
\end{equation*}
Hence, 
\begin{equation*}
\frac{S(X,\mathcal{D}_\mathrm{y}) + S(Y,\mathcal{D}_\mathrm{x})}{S\left(X,\mathcal{D}_\mathrm{x}\right) + S\left(Y,\mathcal{D}_\mathrm{y}\right)}>1\,\ \mathrm{for}\,\ X\neq Y.
\end{equation*}

Intuitively, $d_S$ measures how efficient, on average, is it to approximate one image $X$ using the information of $Y$ extracted in the form of a dictionary of its dominant local structures. The smaller the values of $d_S$ the higher is similarity between the two images.\\

\noindent\textbf{Properties of $d_S$:} 
\begin{itemize}
\item \emph{Non-negativity:} $d_S$ is always non-negative, the lowest value of $d_S$ is $0$ when $X=Y$.
\item \emph{Symmetry:} Clearly, $d_S$ is symmetric i.e.\ $d_S(X,Y) = d_S(Y,X)$. Symmetry is an important property for a similarity or dissimilarity measure because many algorithms (e.g.\ spectral clustering) rely on this property.
\item \emph{Metricity:} $d_S$ does not follow the metric axiom of triangle inequality and hence cannot be called a metric. It would have been mathematically convenient if $d_S$ was a metric. However, many researchers have argued that perceptual distances are typically non-metric in nature \cite{features, emd}.
\end{itemize}

Note that, we have used $p=0$ to compute the complexity functions because our dictionary learning method uses greedy $\ell_0$ approximation. If $\ell_1$ optimization is used to learn the dictionaries, it would be better to use $p=1$ for the definitions. 
\section{Experimental Validation}
\label{sec:experiments}
\begin{figure}[t]
\centering
\begin{tabular}{ccc}
	\includegraphics[width=0.3\linewidth, trim=3cm 3cm 3cm 3cm, clip=true]{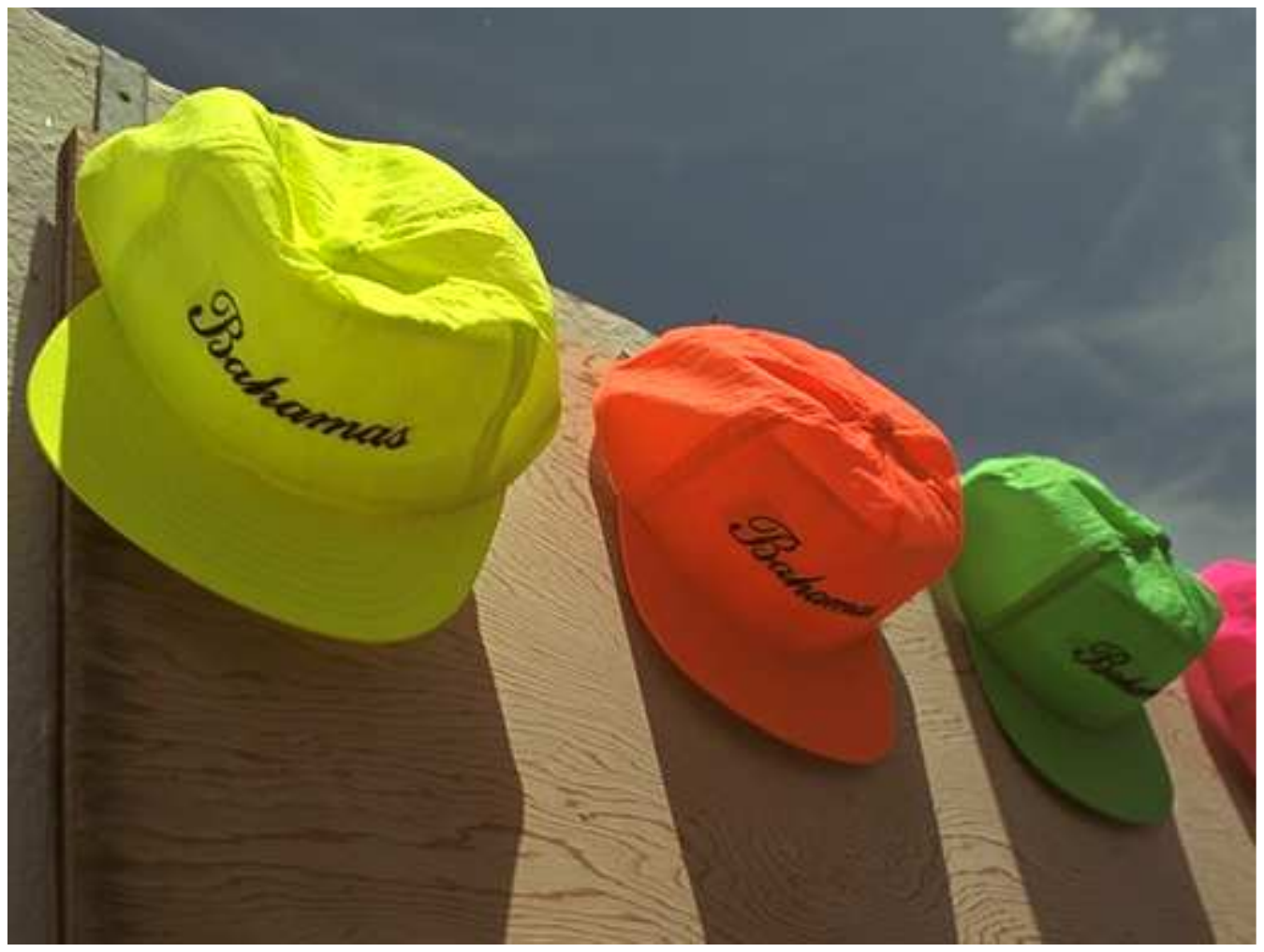} &
	\includegraphics[width=0.3\linewidth, trim=3cm 3cm 3cm 3cm, clip=true]{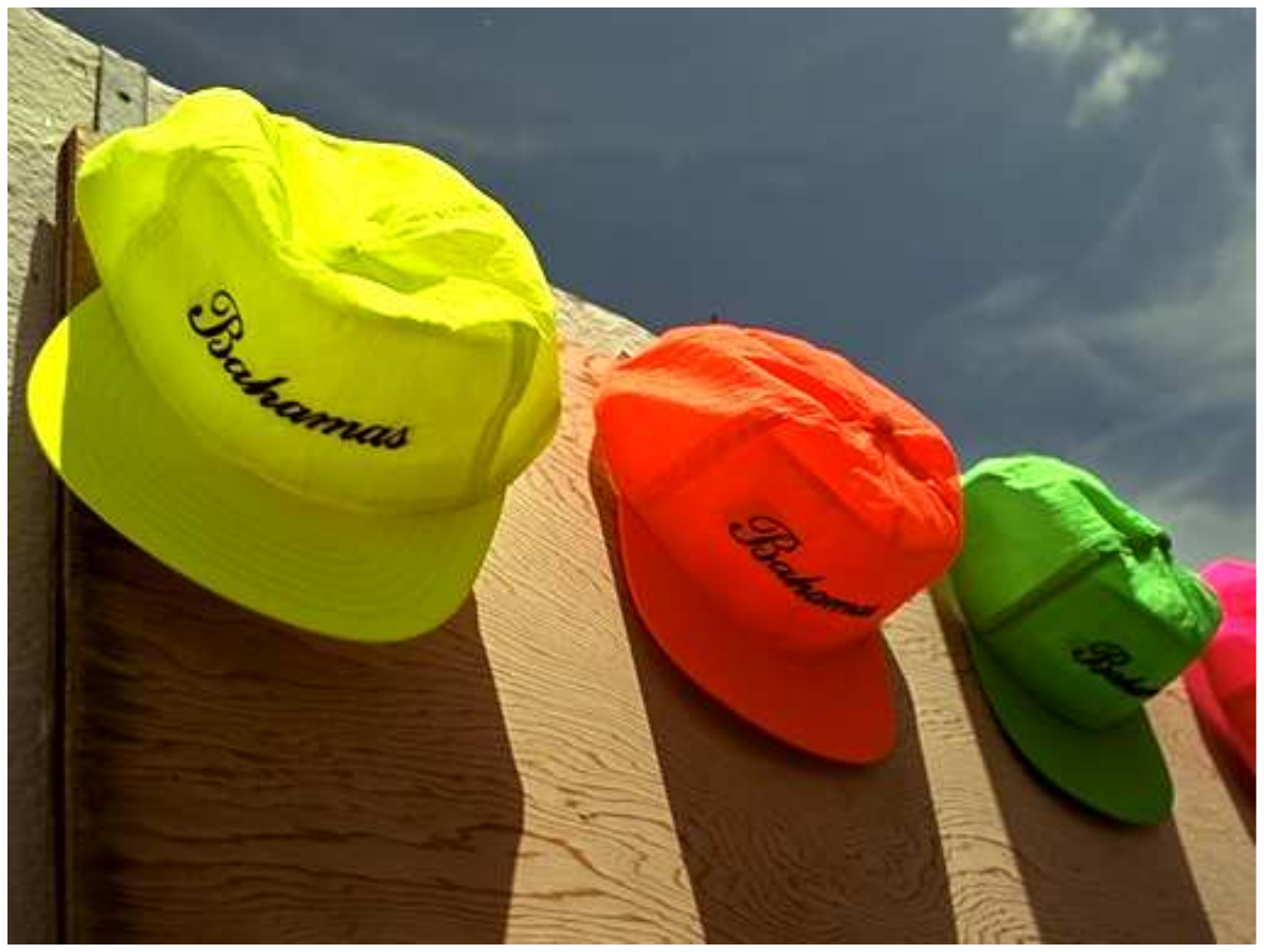} &
	\includegraphics[width=0.3\linewidth, trim=3cm 3cm 3cm 3cm, clip=true]{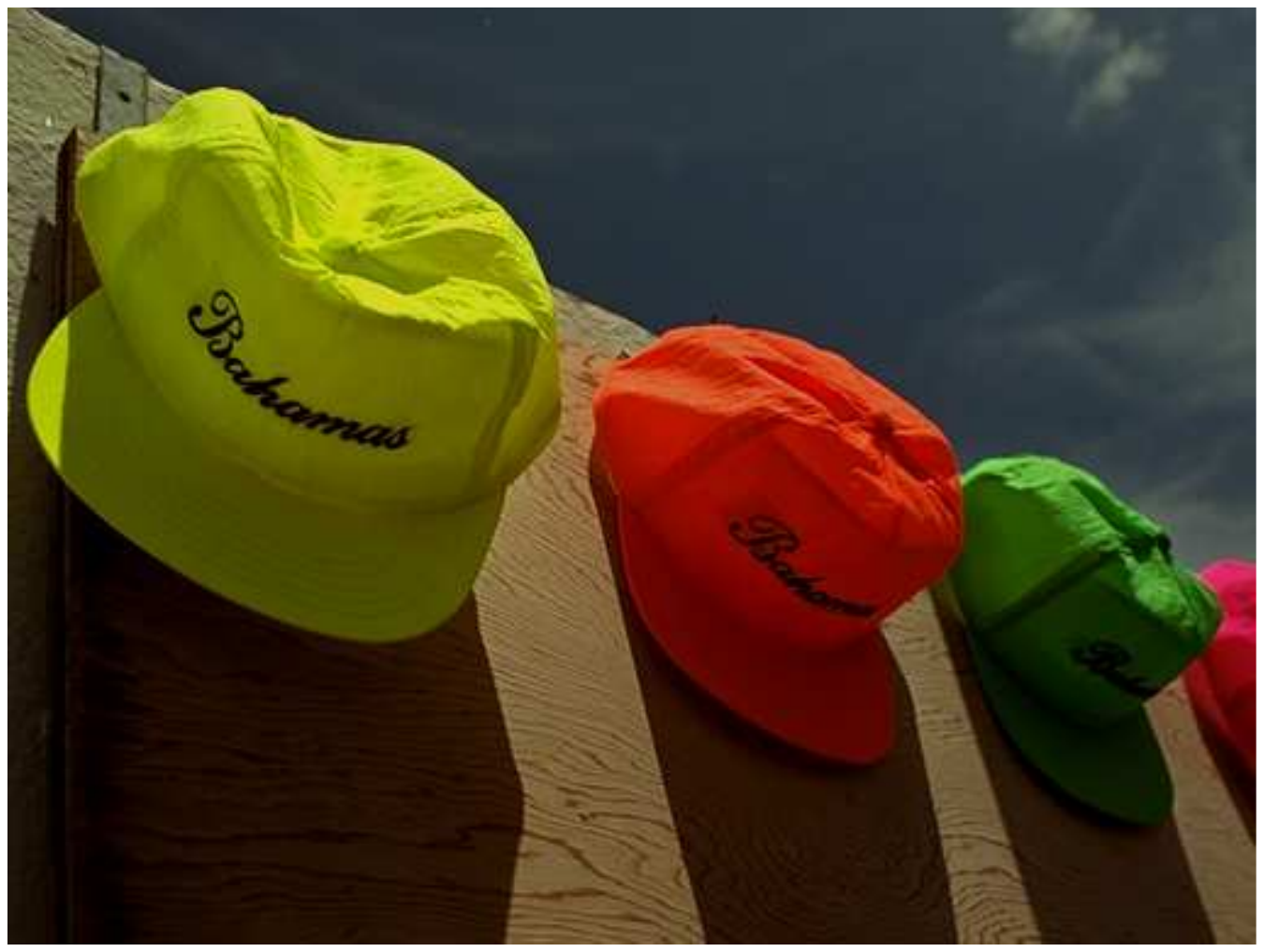} \tabularnewline
	\footnotesize (a) originial image  & \footnotesize (b) contrast change  & \footnotesize (c) luminance change \tabularnewline
	 \footnotesize $\mathrm{PSNR} = \infty$,  $\mathrm{VIF} = 1$ &   \footnotesize $\mathrm{PSNR} = 24.53$, $\mathrm{VIF} = 1.50$ & \footnotesize $\mathrm{PSNR} = 15.97$,  $\mathrm{VIF} = 0.95$\\
	\footnotesize Proposed $ = 0$ &  \footnotesize Proposed $= 0.17$  & \footnotesize Proposed$ = 0.20$\tabularnewline
	
	\includegraphics[width=0.3\linewidth, trim=3cm 3cm 3cm 3cm, clip=true]{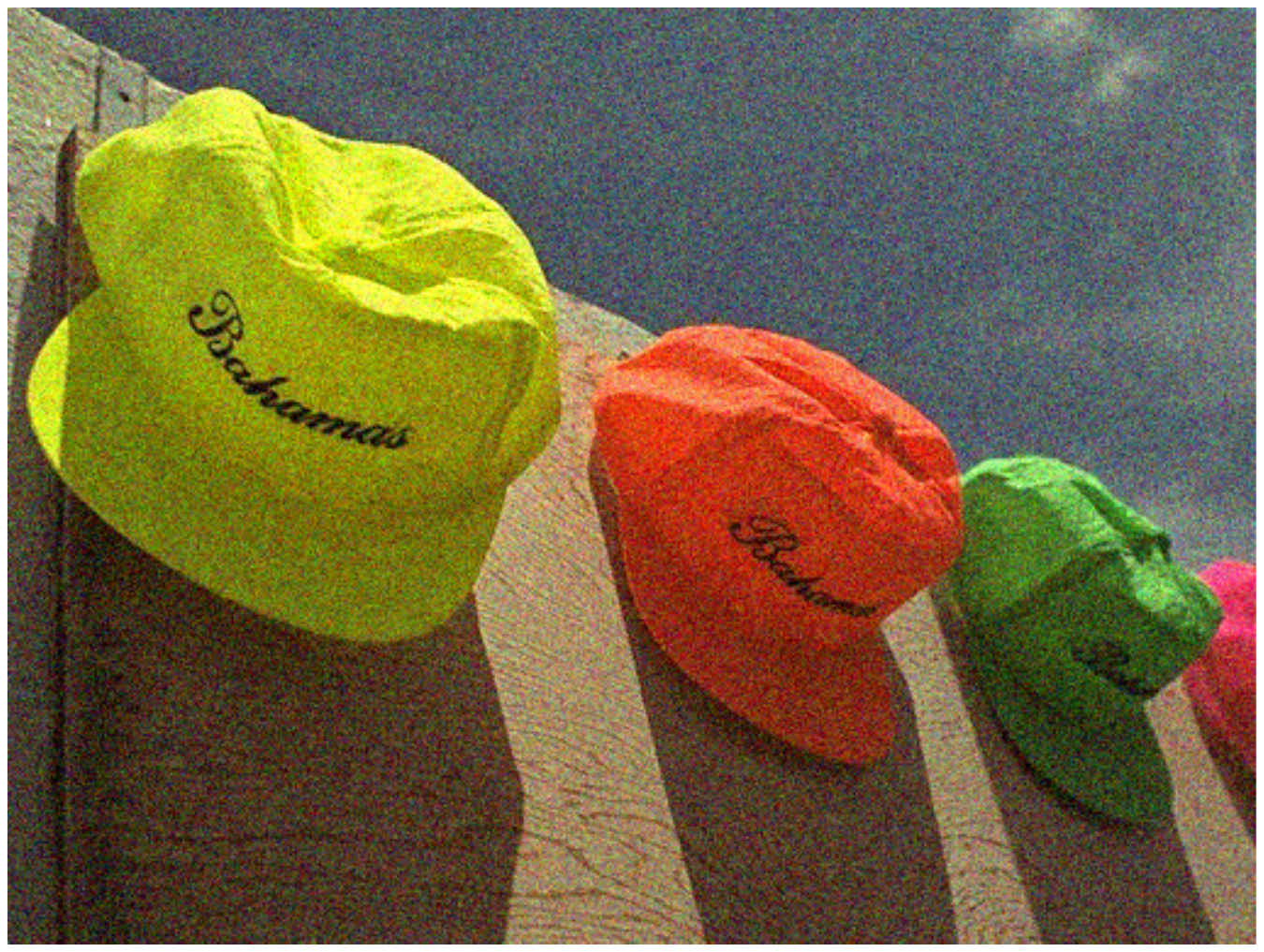} &
	\includegraphics[width=0.3\linewidth, trim=3cm 3cm 3cm 3cm, clip=true]{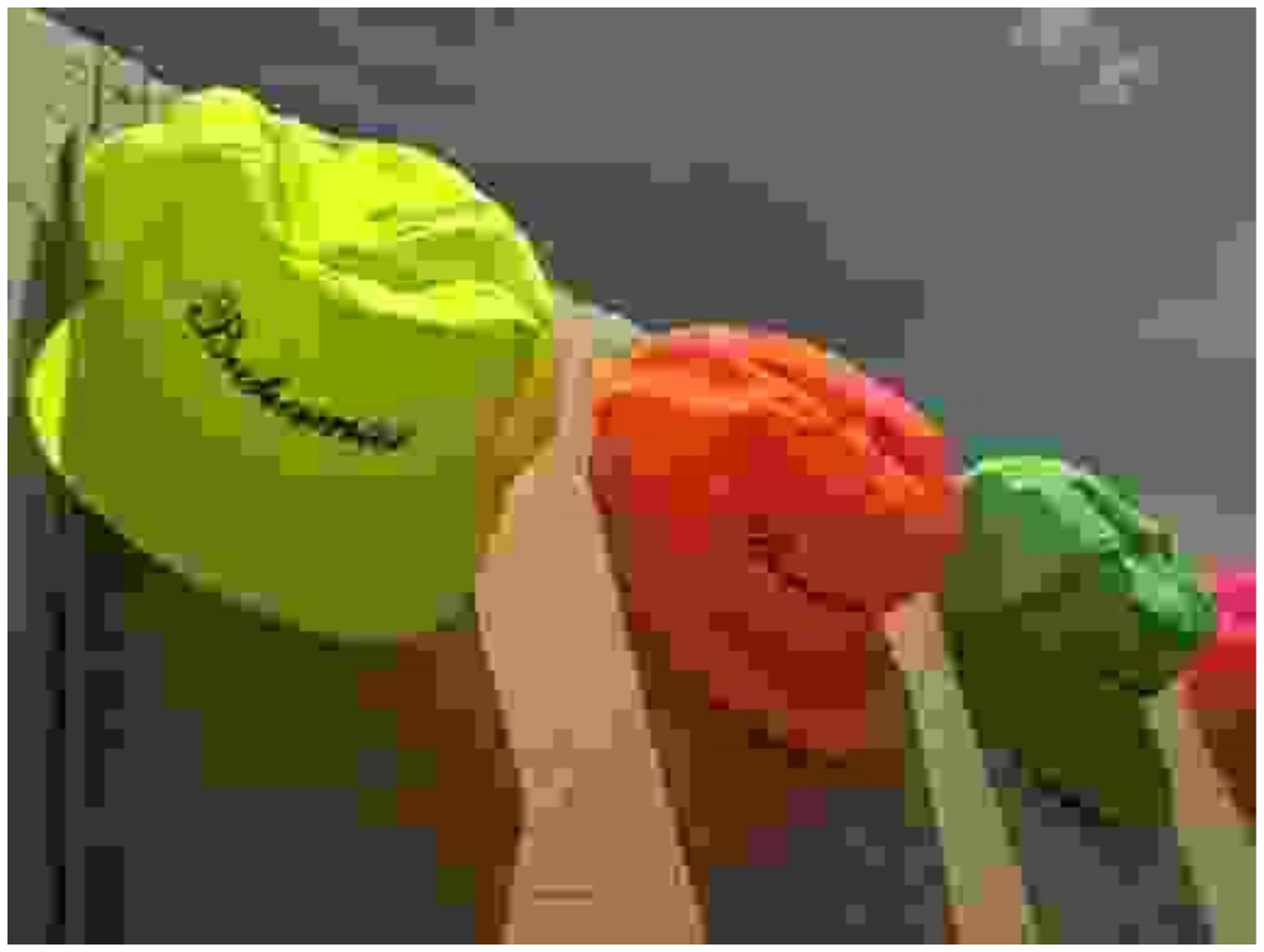} &
	\includegraphics[width=0.3\linewidth, trim=3cm 3cm 3cm 3cm, clip=true]{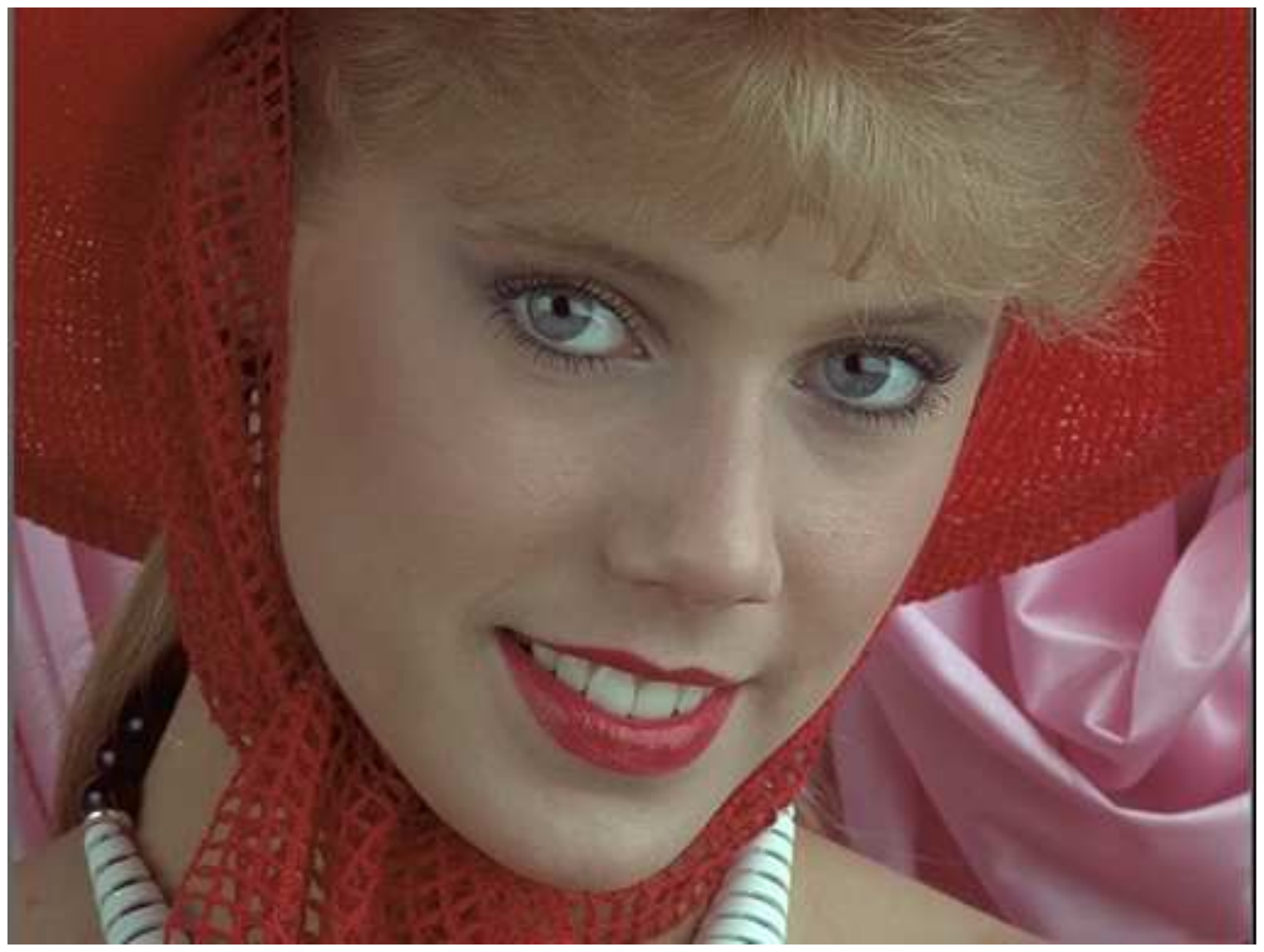} \tabularnewline
	 \footnotesize (d) white noise & \footnotesize (e) lossy jpeg  &  \footnotesize (f) unrelated image  \tabularnewline
	 \footnotesize $\mathrm{PSNR} = 31.95$, $\mathrm{VIF} = 0.96$ & \footnotesize $\mathrm{PSNR} = 28.47$, $\mathrm{VIF} = 0.92$ & \footnotesize $\mathrm{PSNR} = 13.21$, $\mathrm{VIF} = 0.14$\\
	 \footnotesize Proposed$ = 0.33$ & \footnotesize  Proposed$ = 0.38$ &  \footnotesize  Proposed $ = 0.54$\tabularnewline	
\end{tabular}
\caption{The proposed distance measure correlates well with human perception and with the well-known VIF method that measures perceptual signal fidelity.}
\label{fig:quality}
\end{figure}
In order to establish the generality of the proposed distance measure, we perform experiments on a variety of applications. We first perform experiments to evaluate the compatibility of the proposed measure with the human perception of similarity. This is followed by clustering, retrieval and classification experiments involving larger datasets. The datasets that we choose contain real-world images from different domains like biology, biometrics, medicine and natural textures.
\subsection{Implementation Details}
\label{subsec:imple}
Practically, there are  $4$ parameters to be set: the patch size ($\sqrt{m}$), the number of patches to be extracted from each image ($k$), the number of dictionary elements ($n$) and the reconstruction error ($\epsilon$). Unfortunately, there is no theoretical guidelines to determine the values of these parameter, so we rely on previous work and empirical methods. We have used the same parameter values for all experiments, unless mentioned otherwise. Below, we describe how the parameter values are chosen for this particular work.

\emph{Patch size ($\sqrt{m}$) and automatic scale selection:} The patch size determines the spatial scale at which an image is analyzed. For simplicity and speed, we analyze each image at a single scale, but use a simple technique to \emph{automatically} select the (sub)optimal scale. A 2D Laplacian of Gaussian (LoG) filter is applied to each image to detect the local maxima points (keypoitns) at four different scales. The scale at which the maximum number of keypoints are detected is chosen as the (sub)optimal scale for that image. The image is downsampled accordingly and a set of patches are extracted. For example, if the scale is found to be $2$, the image is downsampled by a factor of $2$ and then patches of size $8\times 8$ i.e.\ $\sqrt{m}=8$ are extracted. This particular patch size is chosen in order to be consistent with most of the compression based algorithms (e.g.\ JPEG1) which process $8\times 8$ blocks. The automatic scale selection is performed on all images for all datasets except for the VVT Wood dataset due to the small dimensions ($64\times 64$) of the original images.

\emph{Number of patches ($k$):} In order to train a dictionary, a large number of patches need to be extracted. The color images are first converted to grayscale to achieve \emph{color invariance}. It is also important that the randomly extracted patches contain important structural information of the image and do not come from the homogeneous regions of the image only. This is accomplished by selecting the patches whose energy levels are above an empirically set threshold. A collection of $k = 3000$ such patches are extracted from every image and is used to train its corresponding dictionary. 
The input patches for dictionary learning have zero mean and unit standard deviation which account for \emph{luminance and contrast invariance}.

\emph{Overcompleteness ($n/m$):} Since we intend to learn an overcomplete dictionary, we must have $n>m$. The ratio $n$/$m$ is called the \emph{overcompleteness factor}. It has been shown that for small overcompleteness factor, sparse representation is stable in the presence of noise \cite{wohlberg}. Thus we set $n$/$m = 2$, where $m=64$. 

\emph{Reconstruction error ($\epsilon$):} We used $\epsilon=0.1$ which means that the input vector is reconstructed with at least $90\%$ accuracy. Note that a lower reconstruction error can produce a better dictionary, but requires more computation and more importantly, may cause overfitting.
\begin{figure*}[tb]
	\centering
		\includegraphics[angle=90, width=0.9\linewidth, trim=0cm 0cm 0cm 0cm, clip=true]{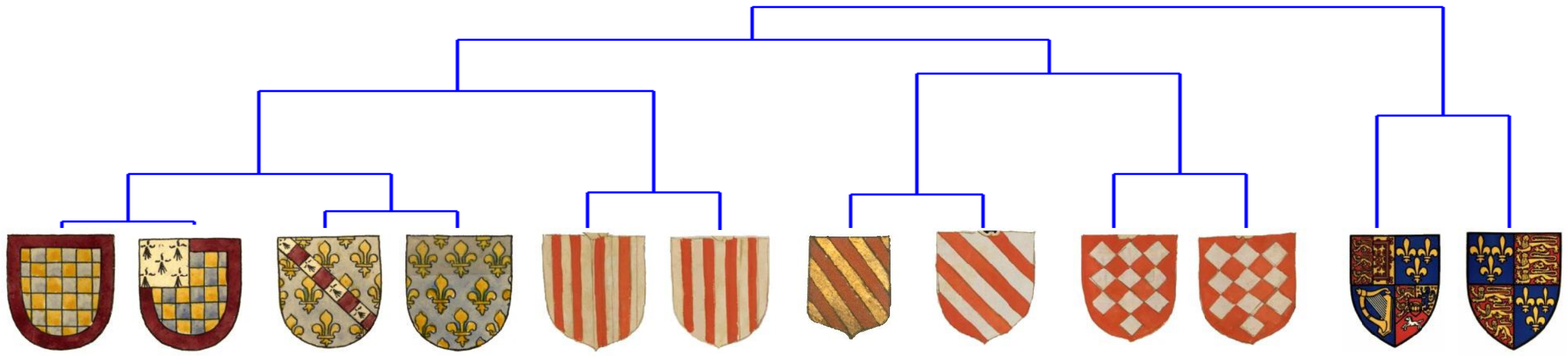}
\caption{Hierarchical clustering result on the Heraldic Shields dataset using the proposed sparse representation-based distance measure (although color images are shown here the result is obtained using grayscale images).}
	\label{fig:clustershield}
\end{figure*}
\subsection{Correlation with Human Perception}
\label{subsec:perception}
It is important that the distance measure between images correlate with human perception. We begin with measuring the similarities between a reference image (Fig.~\ref{fig:quality}(a)) and its distorted versions (Fig.~\ref{fig:quality}(b)-(e)) as well as a completely unrelated image (Fig.~\ref{fig:quality}(f)). We also compare our results with PSNR and the well-known \emph{Visual Information Fidelity} (VIF) \cite{vif} (values closer to zero indicates lower similarity) similarity measure. Figure \ref{fig:quality} shows that the proposed measure correlates well with human perception and with VIF.

Next, we perform a simple clustering task where it is possible to evaluate the results manually. The \emph{Heraldic Shields dataset} \cite{ck} (see Fig.\ \ref{fig:clustershield}) contains $12$ images (of various sizes) which are to be clustered into $6$ pairs. All possible pairwise distances are computed using the proposed distance measure. Hierarchical clustering is performed using the average linkage method. The clustering result shown in Fig.\ \ref{fig:clustershield} demonstrates that our measure has discovered all $6$ basic pairs of shields, and corresponds well with human intuition.
\begin{figure}[tb]
\centering
	\begin {tabular}{c}
		\centerline{\includegraphics[width=0.6\linewidth, trim= 0cm 0cm 0cm 0cm, clip=true]{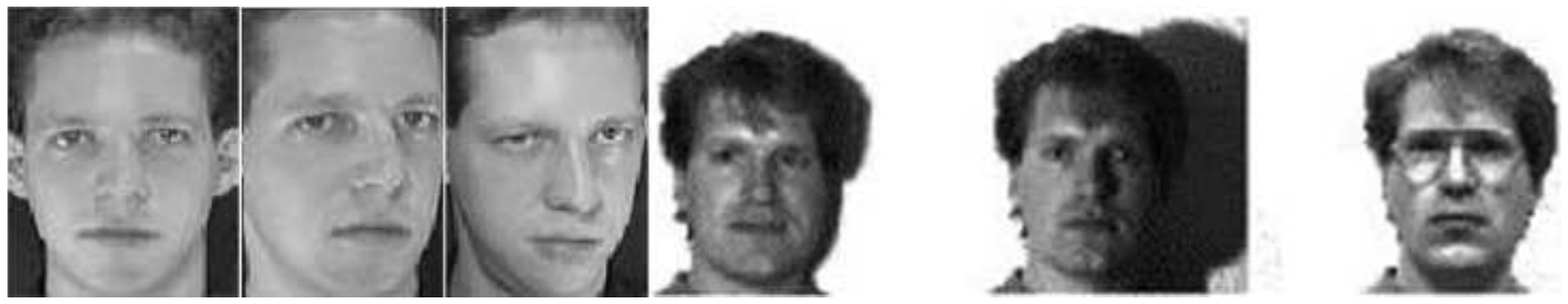}}\\
		\footnotesize (a) \\
		\centerline{\includegraphics[width=0.6\linewidth, trim= 1cm 1.2cm 1cm 1.2cm, clip=true]{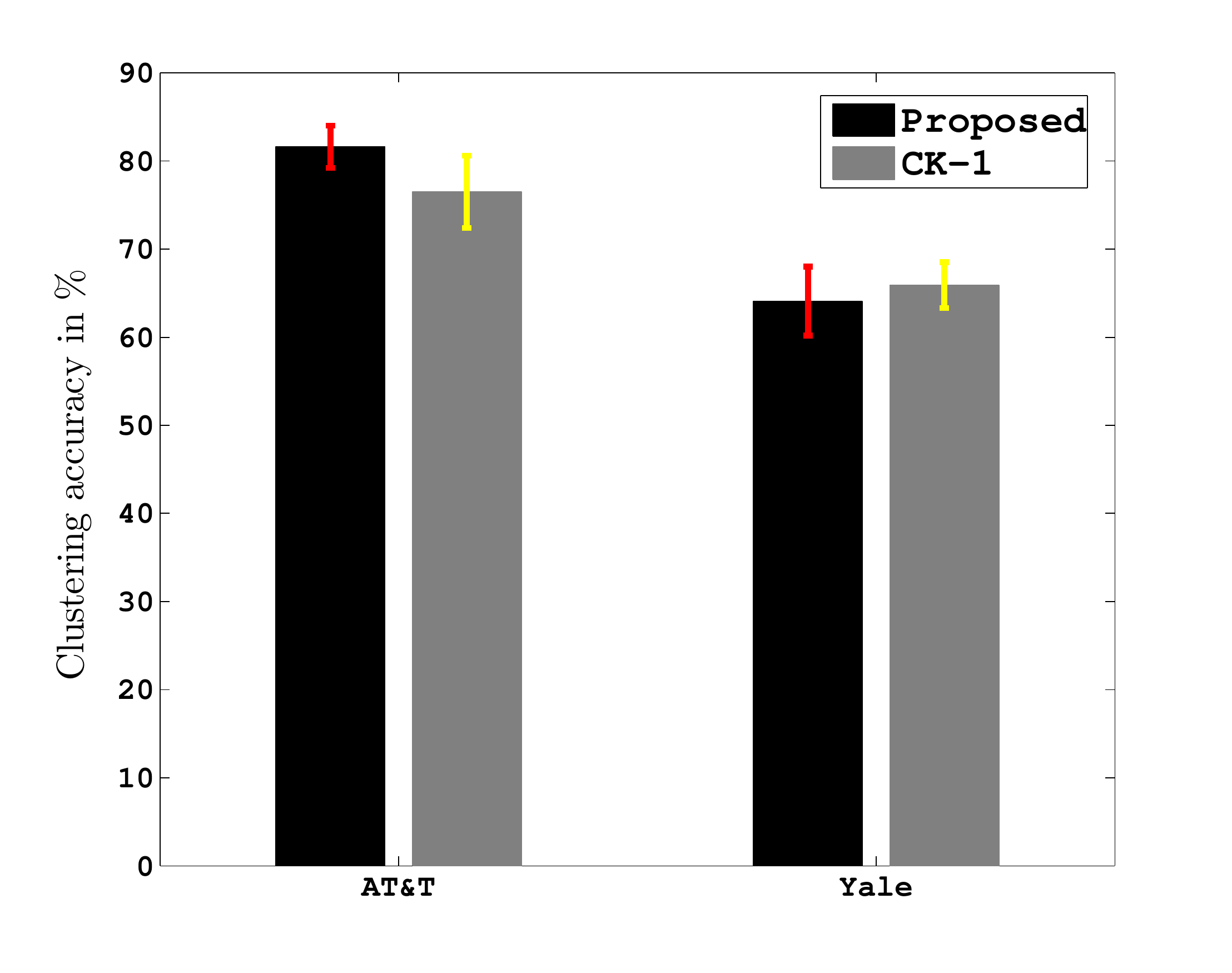}}\\
		\footnotesize(b)
	\end{tabular}
		\caption{(a) Sample images from the AT\&T (first 3) and the Yale face (last 3) databases; (b) Clustering accuracy for the AT\&T face (Proposed: $81.6\pm 2.4\%$, CK-1: $76.5\pm 4.1 \%$) and the Yale face (Proposed: $64.1\pm 3.9\%$, CK-1: $65.9\pm 2.6 \%$) databases.}
	\label{fig:clusterface}
\end{figure}
\subsection{Clustering facial images}
In this segment, we move towards more difficult clustering problems involving two larger benchmark datasets:

\noindent
\emph{AT\&T face} \cite{att}: This dataset contains $400$ facial images of $40$ individuals in $10$ poses. These images (dimension: $112\times 92$) are taken at different times with varying illumination, facial expressions and details. 

\noindent
\emph{Yale face} \cite{yaleface}: This dataset has $165$ grayscale facial images of $15$ individuals. There are $11$ images per subject, one per different condition: center light, with glasses, happy, left light, no glasses, normal, right light, sad, sleepy, surprised, and wink. 

For each dataset, an $M\times M$ similarity matrix is computed using \eqref{eq:sdm}, where $M$ is the number of elements in the dataset. This similarity matrix serves as the input to a standard spectral clustering algorithm \cite{NgSpectral}. The accuracy of the clustering results is measured using the Hungarian algorithm \cite{Hungarian}. We compare our results with the compression-based state-of-the-art CK-1 distance measure \cite{ck} using the code provided by the authors. 
Due to the initialization process in spectral clustering, the accuracy varies slightly at each run. Figure \ref{fig:clusterface} reports the mean clustering accuracies along with the standard deviations as computed over $10$ runs for the two databases under consideration. The proposed measure outperforms CK-1 on the AT\&T face dataset by $5.1\%$ and its performance is $1.8\%$ lower than CK-1 on the Yale dataset.
\subsection{Texture retrieval}
\label{subsec:retrieval}
An image retrieval system, when provided with a query image, returns images from a large dataset that are perceptually similar to the query. We perform standard retrieval experiments on the following benchmark texture dataset.

\emph{Brodatz texture dataset} \cite{brodatz}: This is a benchmark dataset that contains a variety of natural textures like grass and cloth. There are $111$ different texture classes. Each original texture image is divided into $9$ subimages to create the samples for that class. 

For each query, the distances between the query and the remaining $998$ images in the dataset are computed, and the first $K$ nearest images are retrieved. The performance of a retrieval system is often measured in terms \emph{Precision} and \emph{Recall accuracy}. Precision is defined as the ratio of correctly retrieved images to the total number of images retrieved. Recall accuracy is defined as the ratio of the number of correctly retrieved images to the number of images available for the query class. Both precision and recall accuracy are expressed in terms of \%. Our retrieval results are compared with those obtained using the CK-1 method in Fig. {fig:retrv} where our method clearly outperforms CK-1.

\begin{figure}[tb]
\centering
		\includegraphics[width = 0.4\linewidth, trim = 1.5cm 0cm 2cm 1cm, clip = true]{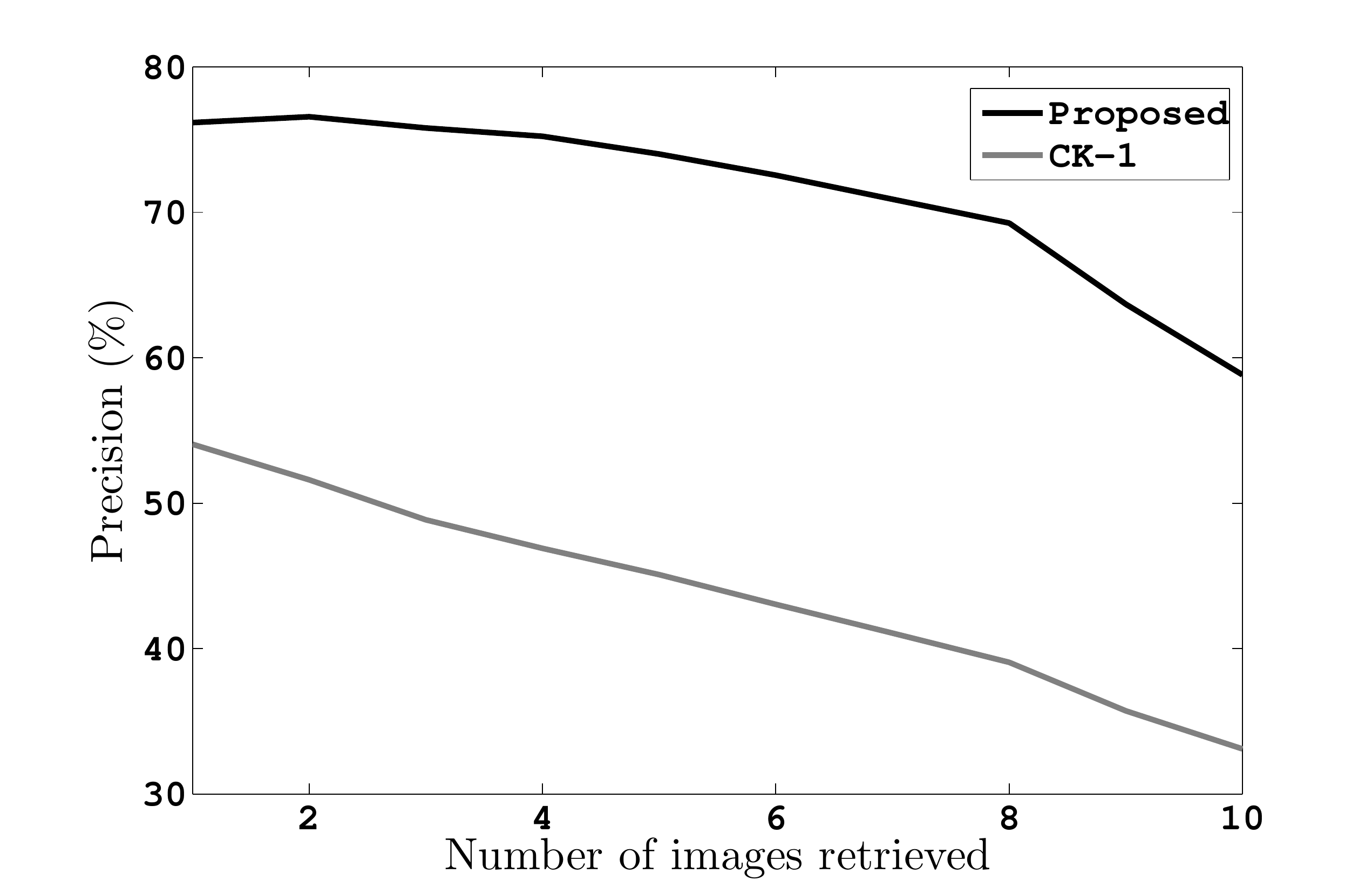}
		\includegraphics[width = 0.4\linewidth, trim = 1.5cm 0cm 2cm 1cm, clip = true]{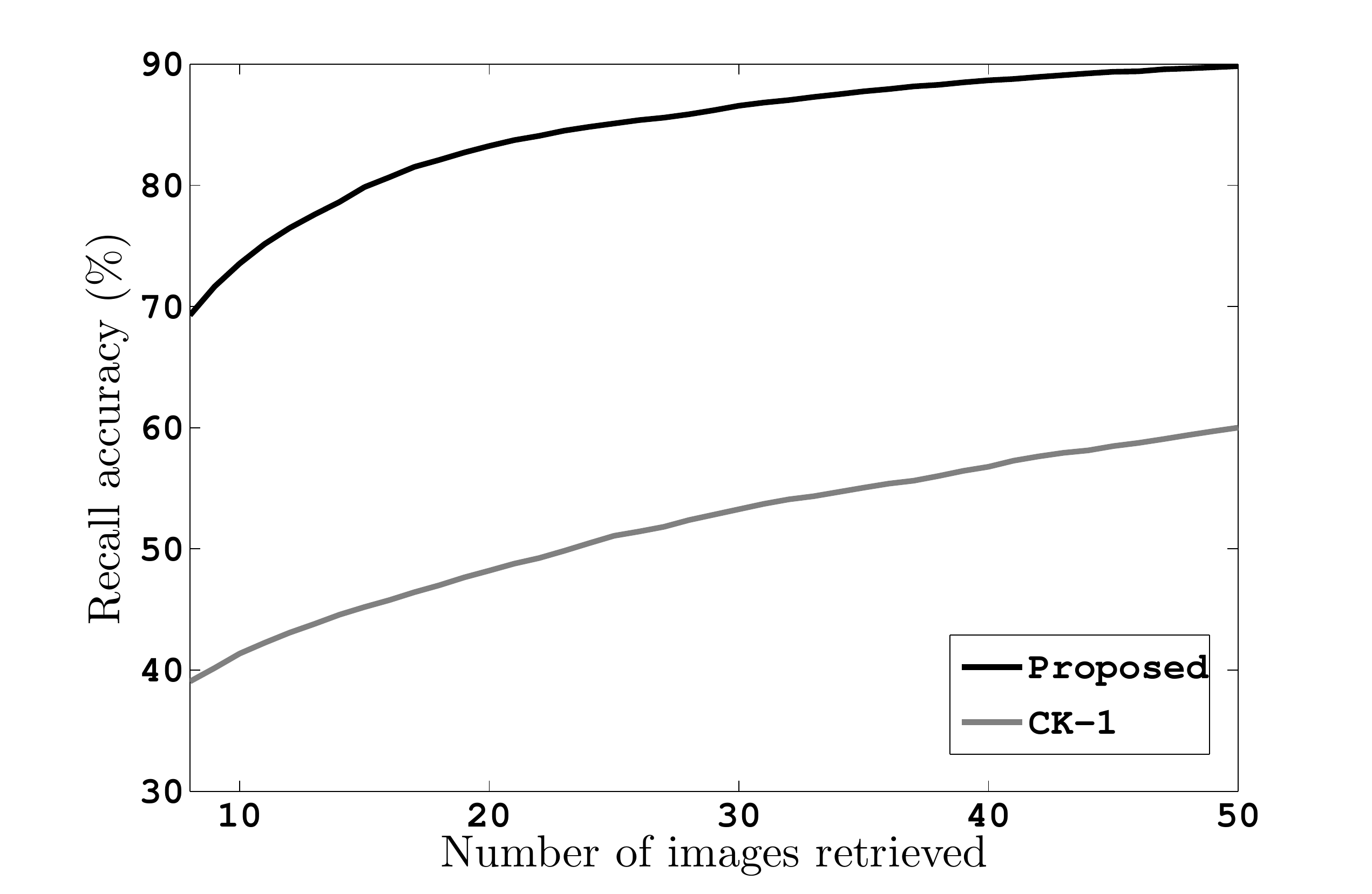}
		\caption{Shown are the image retrieval results in terms of precision (left) and recall accuracy (right) obtained using the proposed method and the compression-based state-of-the-art CK-1 method on the Brodatz dataset.}
	\label{fig:retrv}
\end{figure}
\subsection{Texture classification}
\label{subsec:classification}
Supervised classification experiments are performed on a diverse collection of texture datasets drawn from the sources across various disciplines such as biology, medicine, forensics, etc. 

\noindent\emph{UIUCTex} \cite{uiuc}: This dataset features $25$ texture classes with $40$ samples each. 

\noindent\emph{KTH Tips} \cite{kth}: This dataset consists of textures of $10$ different materials. The images vary in illumination, pose and scale. 

\noindent\emph{Camouflage} \cite{ck}: This dataset consists of $80$ images of $9$ varieties of modern US military camouflage. The images are created by photographing military t-shirts at random orientations.

\noindent\emph{Nematodes} \cite{ck}: Nematodes are wormlike animals with great commercial and medical importance. Their species are often very difficult to distinguish from each other. This dataset contains $50$ images of $5$ different species of nematodes. 

\noindent\emph{Tire tracks} \cite{ck}: This is a collection of tire imprints left on a paper. It has $48$ imprints of $3$ different tires at varying directions.

\noindent\emph{Spiders} \cite{ck}: This is a collection of images of Australasian ground spiders of the family Trochanteriidae. This family has high intra and inter-class variation.

\noindent\emph{VVT Wood} \cite{ck}: This dataset contains $200$ images of $40$ types of wood defects (such as dry knot and small knot, etc.). The task is to label an image as either defective or sound.

The classification results for the above datasets using the proposed method and the CK-1 are presented in Table \ref{tab:classification}. We test both methods using a leave-one-out scheme in a 1-Nearest Neighbor framework. Our method demonstrates much better or comparable accuracy for all the datasets.
%
\begin{table}[tb]
\caption{\small Classification accuracy on various datasets obtained using the proposed distance measure and the state-of-the-art compression-based distance CK-1.}
	\centering
		\begin{tabular}{|c|c|c|c|}
		\hline
		Dataset          & Classes             & Proposed (\%)       			& CK-1 \cite{ck} (\%)  	\tabularnewline
		\hline\hline
		Brodatz          & $111$                & $\mathbf{76.2}$               		& $54.0$          			\tabularnewline
		\hline    
		UIUCTex        & $25$        		    & $\mathbf{51.6}$                    & $51.0$			\tabularnewline
		\hline
		KTH Tips		 & $10$			    & $84.5$               				& $\mathbf{86.0}$			\tabularnewline
		\hline
		Camouflage   & $9$                   & $87.0$      				           & $\mathbf{87.5}$		  \tabularnewline	
		\hline
		Nematodes    & $5$                   & $\mathbf{62.0}$                      & $56.0$	 	    \tabularnewline
		\hline
		Tire tracks      & $3$ 			   & $\mathbf{79.2}$ 		          				 & $\mathbf{79.2}$      \tabularnewline
		\hline
		Spiders			& $3$				  & $70.4$		                           & $\mathbf{96.3}$		\tabularnewline
		\hline	
		VTT wood     & $2$                & $\mathbf{85.2}$      			  & $80.5$	        \tabularnewline
		\hline
		\end{tabular}
	\label{tab:classification}
\end{table}
%
\subsection{Discussion}
\label{subsec: discussion}
Most compression-based methods use an off-the-shelf compressor (data, image or video compressor) and treat the compressor as a black-box. This makes it difficult to understand which part of the compression algorithm actually estimates the complexity of the data or measures the similarity. Consequently, the compression-based methods are difficult to improve upon, unless one wants to delve into the details of the compression algorithms. The proposed method takes a rather direct approach towards the approximation of complexity, and it is easier to understand and improve. Our method can be easily extended to measure the similarity between any type of signals including audio, video and other type of images such as medical images.

The proposed method requires learning a dictionary for each image. The dictionary learning process takes only a few seconds; for example, with the above-mentioned parameter values, a MATLAB implementation takes $\sim2$ secs to learn a dictionary per image (including the patch extraction process) on a standard PC (intel quad @2.67GHz). This is as fast as any standard feature extraction process. However, our method is still slower compared to the compression-based CK1 measure. This can be explained by the fact that the areas of dictionary learning and sparse representation are still in the developing stage. In other words, unlike the standard compression algorithms, the existing algorithms for learning dictionaries or sparse representations are not yet fully optimized for speed or memory. 

We have used a greedy algorithm (OMP) to solve the sparse optimization problems in this work, primarily for speed and simplicity. Better results may be achieved using $\ell_1$ regularized algorithms but at a higher computational cost. The proposed method is also not parameter-free, it requires a few parameters to be set by the user. 
\section{Conclusion}
\label{sec:conclusion}
The main contribution of this work is the introduction of a sparse representation-based approach for computing a generic image similarity measure. The proposed measure has been shown to be successful in classifying, retrieving and clustering a variety of images as it performs consistently at par or better than the state-of-the-art. Nevertheless, the present work is not closed and we hope that this will stimulate interest in the areas of compression or Kolmogorov complexity-based similarity measurement using sparse representation. 

A very recent work has also addressed the problem of similarity measurement using sparse representation of image features \cite{similarityMM}. However, it addresses the problem from a different perspective and does not have any connection with the compression-based or Kolmogorov complexity-based approaches. 

In this work, we have not focused on speeding up the classification, retrieval or the clustering processes since our objective has been to first demonstrate the usefulness and generality of the new distance measure. Future research will focus on using the measure more efficiently to classify and cluster larger datasets. This will require exploiting sophisticated machine learning techniques. Applications can also be extended to problems such as copy detection and data mining. 

\bibliographystyle{IEEEtran}
\bibliography{refs}
\end{document}